\documentclass{article}

\usepackage[preprint]{neurips_2026}
\usepackage{natbib}
\usepackage{enumitem}
\usepackage{amssymb}
\usepackage{amsthm}
\newtheorem{theorem}{Theorem}
\newtheorem{proposition}[theorem]{Proposition}

\newtheorem{definition}[theorem]{Definition}

\usepackage[utf8]{inputenc} 
\usepackage[T1]{fontenc}    
\usepackage{hyperref}       
\usepackage{url}            
\usepackage{booktabs}       
\usepackage{amsfonts}       
\usepackage{nicefrac}       
\usepackage{microtype}      
\usepackage{xcolor}         
\usepackage{amsmath}
\usepackage{listings}
\usepackage[most]{tcolorbox}
\usepackage{xcolor}
\usepackage{array}
\title{Memory Retrieval for Changing Preferences}

\author{
  Yuehan Qin, Li Li, Linxin Song, Wei Yang, Jiate Li, Yuqing Yang, Yue Zhao\\
  University of Southern California\\
  \texttt{\{yuehanqi@usc.edu\}}
}

\begin{document}

\maketitle

\begin{abstract}
Long-context dialogue systems must decide both when to access memory and which parts of the interaction history are relevant. Existing approaches typically rely on heuristic retrieval signals or always-on memory usage, failing to account for the changing and potentially inconsistent nature of user preferences.
In this work, we propose a unified framework for memory access and selection based on changing preferences. We formulate personalized memory retrieval as identifying which historical turns provide evidence about a user’s latent preference state, rather than relying on surface-level semantic similarity.
To this end, we quantify the utility of each memory turn using a Bayes factor, defined as the improvement in the model’s likelihood of the reference response when the turn is included in context.
This provides a principled measure of evidence strength and a unified signal for both memory access and selection.
By framing memory retrieval as utility estimation, the model learns to identify salient turns and regulate memory usage based on expected utility.
Experiments on four heterogeneous memory benchmarks show that our approach
outperforms existing embedding-based retrieval
on long-context, preference-intensive tasks where modeling changing preferences is essential, while remaining competitive in low-density regimes where semantic similarity suffices.
\end{abstract}

\section{Introduction}
As users engage with conversational AI systems across extended periods, they accumulate histories rich with preferences, experiences, and changing beliefs. A system that cannot selectively draw on this history when answering new queries fails to deliver the kind of continuity that personalization demands \citep{yang2026selfevolvingllmmemoryextraction}.
However, as interaction histories grow to span hundreds of sessions, attending to everything is computationally intractable \citep{ding2024longropeextendingllmcontext, 
li2025personalizedconversationalbenchmarksimulating}, and attending to the wrong things actively degrades response quality \citep{liu2023lostmiddlelanguagemodels,shi2023largelanguagemodelseasily, Amiraz_2025}. This challenge is further exacerbated by the well-documented tendency of large models to hallucinate or rely on spurious cues when presented with noisy or misleading context~\citep{li-etal-2025-treble,li2026autonomytaxdefensetraining}.

\begin{figure}[t]
    \centering
\includegraphics[width=\columnwidth]{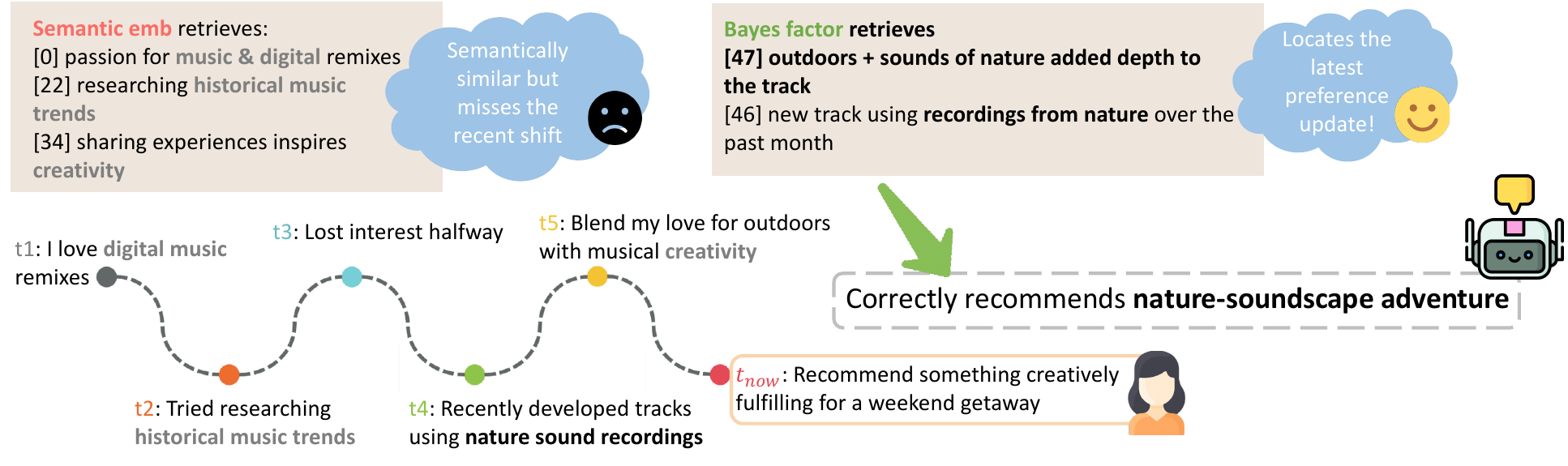}
    \caption{Illustration of retrieval methods for personalization memory. Given a generic recommendation 
request, semantic embedding retrieval returns surface-similar but evidentially uninformative 
turns (e.g., abstract self-reflections about creativity), leading the LLM to a generic 
recommendation. Our Bayes Factor-based retrieval instead locates the user's recent preference 
shift toward nature recording, enabling the LLM to produce the correct preference-aligned 
recommendation.}
    \label{fig:demo}
\end{figure}

This makes memory retrieval a central challenge in personalized AI. Given a new query, the system must identify which past interactions carry genuine utility for producing a correct and contextually appropriate response.
The approach that has emerged as standard practice is semantic embedding similarity: each memory turn and the incoming query are encoded, and the turns nearest in embedding space are selected as context \citep{chhikara2025mem0buildingproductionreadyai,wu2025longmemevalbenchmarkingchatassistants,Li_2025_CVPR,xu2025amemagenticmemoryllm, tan-etal-2025-prospect}.
Semantic similarity is, however, a proxy for the wrong quantity. It measures topical co-occurrence in embedding space, not whether a memory turn provides incremental evidence that is useful for resolving the current query. This distinction is inconsequential in many retrieval settings, but it becomes critical in personalization, where the same topic can appear across turns that reflect fundamentally different states of the user. 
This mismatch manifests in several structural failure modes. First, semantically similar turns may correspond to incompatible or superseded latent preference states, causing retrieval systems to surface context that is topically related yet no longer useful for the current query~\citep{jiang2025knowmerespondme,zhao2025llmsrecognizepreferencesevaluating}. Second, evidential utility is often sparse, temporally distant, or only weakly aligned with the query surface form. As a result, retrieval methods grounded primarily in semantic similarity tend to favor lexical or topical overlap over genuinely decision-relevant evidence. Similar biases toward surface-level correlations have also been observed in other structured prediction settings~\citep{limm,Li_Ji_Wu_Li_Qin_Wei_Zimmermann_2024}. An example is shown in Fig.~\ref{fig:demo}.

This failure is symptomatic of a deeper issue in how existing memory systems decompose the retrieval problem. Most systems treat two coupled decisions independently: whether external memory should influence the response for a given query, and if so, which turns are worth retrieving \citep{packer2024memgptllmsoperatingsystems, chhikara2025mem0buildingproductionreadyai,tan-etal-2025-prospect}. Access is typically handled through always-on or coarse heuristics, while selection relies on semantic similarity\citep{li2025helloagainllmpoweredpersonalized, wu2025longmemevalbenchmarkingchatassistants}. Neither decision is grounded in a direct estimate of memory utility for the current query. The result is a system that retrieves what is related rather than what is useful.

We argue that both decisions can be unified under a single principle: a memory turn should be retrieved if and only if it provides evidence that improves the correctness of the response, and memory should be accessed at all only when such evidence exists. We operationalize this principle through a Bayesian lens, treating each candidate turn as potential evidence and measuring how much it updates the model's belief toward the correct answer.

We summarize our contributions as below:
\begin{itemize}[leftmargin=*]
\item \textbf{A principled retrieval signal for personalized memory.} We interpret utility gain as a Bayes factor, providing a principled alternative to semantic similarity that captures evidential utility rather than topical relatedness.
\item \textbf{Bayes-factor-gated memory framework.} We propose a simple framework that couples (i) a discrete gating mechanism for deciding when to trigger memory access with (ii) a saliency-based selection mechanism for identifying relevant context, both driven by the same Bayes factor signal.
\item \textbf{Evidence of distinctiveness.} Through controlled ablations, we show that Bayes-factor-based saliency labels are not interchangeable with embedding-based alternatives, with the largest gap on long-context preference tasks (+8.76pp on MemBench-High), and that supervised fine-tuning on these labels is essential for the observed gains.
\item \textbf{Improved long-context personalization.} We demonstrate consistent improvements across multiple personalization benchmarks, with the largest gains on challenging instances requiring long-range memory reasoning.
\end{itemize}

\section{Related Works}
Large Language Models have emerged as personalized assistants across a wide range of tasks, from writing support to tailored recommendations~\citep{salemi2024lamplargelanguagemodels,yang2025toward,li2025personalizedconversationalbenchmarksimulating,gu2025r,zhao2025hierarchical,li2025secureondevicevideoood}. As interaction histories accumulate across sessions, effectively leveraging them requires selective memory access: attending to everything is computationally intractable, while attending to the wrong turns degrades response quality~\citep{liu2023lostmiddlelanguagemodels}.
Early memory systems such as MemGPT~\cite{packer2024memgptllmsoperatingsystems} and MemoryBank~\cite{zhong2023memorybankenhancinglargelanguage} address context window limitations by maintaining external memory stores. More recent systems including Mem0~\cite{chhikara2025mem0buildingproductionreadyai} and A-MEM~\cite{xu2025amemagenticmemoryllm} extract and consolidate facts from dialogue history, while RMM~\cite{tan-etal-2025-prospect} introduces reflective summarization for finer retrieval granularity. These systems uniformly rely on semantic embedding similarity for retrieval, which we identify as a structural limitation under preference evolution.

Several works have proposed using utility gain as a context quality signal. RECOMP~\cite{xu2023recompimprovingretrievalaugmentedlms} uses generator feedback for context compression, CODEFILTER~\cite{li2025impact} applies likelihood-based impact scoring to code completion, and~\citet{Liu_2025} show that pointwise mutual information between context and query correlates with answer accuracy. Our work adapts this family of signals to personalized memory retrieval, where its interpretation as a Bayes factor over changing user preferences provides principled motivation beyond prior heuristic applications.

\section{Methodology}
\paragraph{Problem Setup}\label{sec:setup}
Let a dialogue history be $\mathcal{H} = (c_1, c_2, \ldots, c_T)$, a user query $q$, and the (unknown) answer $a$. The goal is to select a subset of turns $F \subseteq \mathcal{H}$ such that a model conditioned on $(q, F)$ can produce the correct answer. This problem involves two coupled decisions:
(i)~\emph{whether} memory should be consulted, and
(ii)~\emph{which} turns should be retrieved.

\subsection{Stage 1: Answer-Conditioned Saliency}
\label{sec:sala}
Following the Bayesian view of evidence, we treat
$\mathrm{Sal}_a(c_i)$ as the \emph{log Bayes factor} of turn $c_i$
for answer $a$ conditioned on query $q$: a turn constitutes genuine
memory evidence if and only if observing it updates the answer
distribution beyond what the query alone establishes.
Formally, $p(a \mid q)$ is the model's prior belief over the answer
given the query alone, and $p(a \mid q, c_i)$ is the posterior after
observing turn $c_i$. Then $\mathrm{Sal}_a(c_i)$ is the log-ratio
between the two:
\begin{equation}
\mathrm{Sal}_a(c_i \mid q, a)
=
\log p(a \mid q, c_i)
-
\log p(a \mid q),
\end{equation} 
which corresponds to the classical log Bayes factor quantifying how strongly $c_i$ supports $a$. In practice, we calculate 
both terms from a smaller language model for computing efficiency.

A turn whose Bayes factor is near unity (e.g., an occupational fact
when the query asks for a location) is correctly suppressed; a turn
that drives the posterior sharply toward the correct answer receives
high $\mathrm{Sal}_a$. This stands in direct contrast to
embedding-based retrieval, which measures query--turn similarity and
thus cannot distinguish \emph{evidential relevance} from
\emph{topical relatedness}.

\textbf{Oracle Focus Set.}
We define the oracle focus set at threshold $\tau$ as
\begin{equation}
S_\tau
\;=\;
\bigl\{
c_i \in \mathcal{H}
:\;
\mathrm{Sal}(c_i \mid q, a) > \tau
\bigr\},
\label{eq:oracle_focus}
\end{equation}
where $\mathrm{Sal}(c_i\mid q,a)$ denotes the answer-conditioned
saliency of turn $c_i$ given query $q$ and gold answer $a$, and
$\tau$ is a threshold controlling the amount of retained evidence.
$S_\tau$ represents an oracle notion of memory: it contains exactly those turns that improve answer prediction.

\subsection{Stage 2: Query-Conditioned Proxy}
\label{sec:salq}

$\mathrm{Sal}_a(c_i; q)$ depends on the unknown gold answer $a$, and is therefore not directly available at inference time. To obtain a tractable approximation, we replace the answer $a$ with the observable query $q$, yielding a query-conditioned Bayes factor:
\begin{equation}
\mathrm{Sal}_q(c_i; q) \;=\; \log p(q \mid c_i) \;-\; \log p(q).
\label{eq:salq}
\end{equation}
Here $p(q)$ is the model's prior probability of the query in isolation, and $p(q \mid c_i)$ is the likelihood of the query after observing turn $c_i$. $\mathrm{Sal}_q$ is therefore the log ratio of the two---a Bayes factor that measures how strongly $c_i$ constitutes evidence for the query\footnote{Equivalently, $\mathrm{Sal}_q(c_i; q) = \mathrm{PMI}(q; c_i)$, connecting our signal to the family of pointwise mutual information context-quality measures studied in retrieval-augmented generation~\citep{Liu_2025}}.

While $\mathrm{Sal}_q$ does not condition on the answer, we now show that it is a principled surrogate for $\mathrm{Sal}_a$ rather than a heuristic substitute. Our argument proceeds in three steps: an exact decomposition (Proposition~\ref{prop:decomp}), a sufficient condition for rank-equivalence (Theorem~\ref{thm:rank}), and a bounded approximation result for the realistic case (Theorem~\ref{thm:bound}).

\paragraph{An exact decomposition.}
By the chain rule of Bayesian updating, the joint Bayes factor of $c_i$ for the pair $(a, q)$ decomposes additively:

\begin{proposition}[Joint Bayes factor decomposition]
\label{prop:decomp}
For any candidate turn $c_i$, query $q$, and answer $a$,
\begin{equation}
\underbrace{\log \frac{p(a, q \mid c_i)}{p(a, q)}}_{\mathrm{BF}_{(a,q)}(c_i)}
\;=\;
\underbrace{\log \frac{p(a \mid q, c_i)}{p(a \mid q)}}_{\mathrm{Sal}_a(c_i)}
\;+\;
\underbrace{\log \frac{p(q \mid c_i)}{p(q)}}_{\mathrm{Sal}_q(c_i)}.
\label{eq:decomp}
\end{equation}
\end{proposition}

\noindent The identity (\ref{eq:decomp}) is exact and assumption-free. It shows that $\mathrm{Sal}_q$ is not an unrelated heuristic but rather one of two additive components of the joint evidence $c_i$ provides for $(a, q)$;
Ranking by $\mathrm{Sal}_q$ is therefore equivalent to ranking by $\mathrm{BF}_{(a,q)} - \mathrm{Sal}_a$, and the question of when this ranking agrees with $\mathrm{Sal}_a$ reduces to a question about the relative magnitudes of the two terms.

\paragraph{When does the surrogate preserve the ranking?}
We now identify a condition under which $\mathrm{Sal}_q$ and $\mathrm{Sal}_a$ induce the same ordering on candidate turns up to a constant.

\begin{definition}[Query-conditional sufficiency]
\label{def:suff}
A turn $c_i$ is \emph{query-sufficient} for $(q, a)$ if $q \perp a \mid c_i$, i.e.,
\begin{equation}
p(a \mid q, c_i) \;=\; p(a \mid c_i).
\label{eq:suff}
\end{equation}
\end{definition}

\noindent Intuitively, $c_i$ is query-sufficient when it already encodes the user-state information needed to determine $a$, so that the query merely \emph{triggers} retrieval rather than supplying additional evidence. This holds naturally for the kind of turns personalized memory aims to retrieve: declarative preference statements, persona attributes, and committed user states.

\begin{theorem}[Rank-equivalence under sufficiency]
\label{thm:rank}
Let $\mathcal{C} = \{c_1, \ldots, c_n\}$ be a candidate set in which every turn is query-sufficient (Def.~\ref{def:suff}). Then for all $c_i, c_j \in \mathcal{C}$,
\begin{equation}
\mathrm{Sal}_a(c_i) - \mathrm{Sal}_a(c_j)
\;=\;
\mathrm{Sal}_q(c_i) - \mathrm{Sal}_q(c_j)
\;+\;
\bigl[\mathrm{Sal}_a^{\mathrm{marg}}(c_i) - \mathrm{Sal}_a^{\mathrm{marg}}(c_j)\bigr],
\label{eq:rank}
\end{equation}
where $\mathrm{Sal}_a^{\mathrm{marg}}(c_i) = \log p(a \mid c_i) - \log p(a)$ is the marginal answer Bayes factor. In particular, if $\mathrm{Sal}_a^{\mathrm{marg}}$ is approximately constant across $\mathcal{C}$, then ranking by $\mathrm{Sal}_q$ recovers the $\mathrm{Sal}_a$ ranking exactly.
\end{theorem}

\noindent The constant-$\mathrm{Sal}_a^{\mathrm{marg}}$ assumption is mild whenever the candidate pool has been pre-filtered to a topically coherent slice of the user's history, which is precisely what the query-side top-$K$ pre-filter in our inference pipeline (Eq.~\ref{eq:prefilter}) accomplishes.

\paragraph{The realistic case.}
Strict query-conditional sufficiency rarely holds, but the disagreement between $\mathrm{Sal}_q$- and $\mathrm{Sal}_a$-induced rankings can be bounded in terms of a per-turn sufficiency residual $\varepsilon_i := |\log p(a \mid q, c_i) - \log p(a \mid c_i)|$ and the marginal-evidence gap. We state and prove the precise bound in Appx.~\ref{app:proofs} (Theorem~\ref{thm:bound}); the practical takeaway is that $\mathrm{Sal}_q$ is faithful when turns are self-contained and degrades on instances requiring cross-turn aggregation.

\subsection{Stage 3: Threshold-Gated Memory}
\label{sec:threshold}
\begin{figure}[t]
    \centering
    \includegraphics[width=\columnwidth]{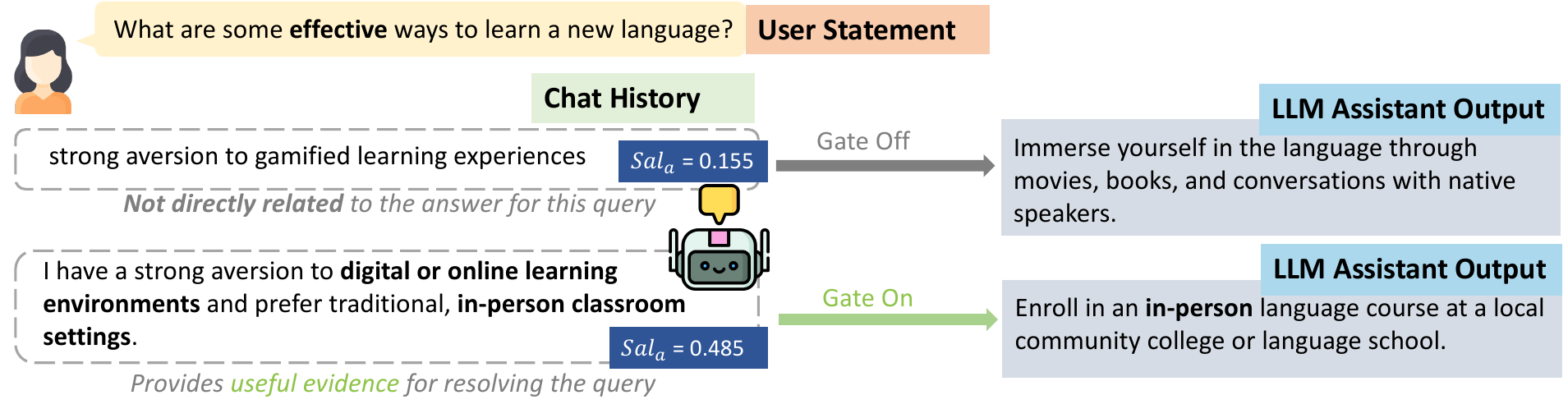}
    \caption{
Illustration of threshold-gated memory access. Although both memory turns are semantically related to the user query, only the lower example provides evidence that materially changes the model’s belief about the appropriate response, resulting in a higher $\mathrm{Sal}_a$ score
and activation of memory-guided reasoning.
}
    \label{fig:gate}
\end{figure}
We convert the continuous saliency signal into a discrete decision via a threshold $\tau$:
\begin{equation}
S_\tau(\mathcal{H}, q)
=
\{\, c_i \in \mathcal{H} : \mathrm{Sal}_q(c_i; q) > \tau \,\},
\quad
g_\tau(\mathcal{H}, q)
=
\mathbf{1}\!\left[\,|S_\tau(\mathcal{H}, q)| > 0\,\right].
\label{eq:gate}
\end{equation}

Here $S_\tau$ is the set of \emph{salient chunks} retained as evidence,
and $g_\tau\!\in\!\{0,1\}$ is the memory \textbf{trigger}
indicating whether the memory-guided policy should fire on this
context.

\textbf{Why couple $\mathrm{Sal}_a$ and $\tau$?}
$\mathrm{Sal}_a$ alone provides a continuous utility score but no
decision boundary, so the model cannot learn a discrete retrieval policy.
A fixed threshold alone provides a gate but no ranking, so the model
cannot prioritize among retained turns. Coupling them gives a
calibrated labeling mechanism that (i)~filters low-utility turns via
$\tau$, (ii)~ranks the survivors by their Bayes factor score, and
(iii)~exposes both the relevance label and the ranking as explicit
supervision targets for the controller. This is the central design
choice of our silver label construction.

\paragraph{Implications for the pipeline.}
Theorem~\ref{thm:rank} delineates the regime in which $\mathrm{Sal}_q$ is a faithful surrogate: candidate sets dominated by self-contained user-state turns. Outside this regime where instances requiring cross-turn aggregation, the supervised controller (\S~\ref{sec:train_pipeline}), trained on $\mathrm{Sal}_a$ silver labels, recovers ranking quality that $\mathrm{Sal}_q$ alone cannot. The two-stage design ($\mathrm{Sal}_q$ pre-filter $\rightarrow$ $\mathrm{Sal}_a$-trained controller) is therefore a direct response to this structure.

\subsection{Stage 4: Learning the Retrieval Policy}
\label{sec:learning}

\subsubsection{Training pipeline.}\label{sec:train_pipeline}
The trigger $g_\tau$ and the salient set $S_\tau$ are exposed to a
\emph{single} fine-tuned policy through two paired supervised signals
constructed from the same $(\mathcal{C}, S_\tau)$. We instantiate them
in three stages, all sharing parameters under one
next-token-prediction objective.

\smallskip
\textbf{Controller head (memory-gating).}
Given $\tau$, the user query $q$, and the candidate evidence
$S_\tau$, the model is trained to emit a structured action
\begin{equation}
a \;=\; \big(
\mathrm{trigger}=g_\tau,\;
\mathrm{focus\_idx}=\mathrm{TopK}_{\mathrm{Sal}_a}(S_\tau),\;
\mathrm{policy}=\pi^\star
\big),
\label{eq:controller}
\end{equation}
wrapped in a brief natural-language $\langle\mathrm{thought}\rangle$
followed by a JSON $\langle\mathrm{action}\rangle$. Here
$\mathrm{TopK}_{\mathrm{Sal}_a}$ ranks the candidate chunks by their
answer-conditioned saliency score. To prevent the controller from
collapsing to $g_\tau\!\equiv\!1$, we balance with negative samples (details in Appx.~\ref{app:implement}).

\smallskip
\textbf{Reader head (MCQ answering).}
Conditioned on the same $S_\tau$ and the four candidate options
$\mathcal{O}$, the model is trained to produce the gold answer
terminated by $\langle\mathrm{final\_answer}\rangle(x)$.

\smallskip
\textbf{Joint objective.}
Both heads share parameters and are trained jointly with a single
next-token-prediction objective on the union of controller and reader
samples:
\begin{equation}
\mathcal{L}_{\text{SFT}} \;=\;
-\!\!\sum_{(x,y)\in\mathcal{D}_{\text{ctrl}}\cup\mathcal{D}_{\text{read}}}
\sum_{t}\log p_{\theta}\!\bigl(y_t\mid y_{<t},x\bigr).
\label{eq:loss}
\end{equation}

\subsection{Inference Pipeline}
During inference, we have a query $q$ together with the user's full
conversation history $\mathcal{H}=\{c_1,\dots,c_n\}$ up to $q$, but neither the
gold answer $a$ nor the answer-conditioned saliency
$\mathrm{Sal}(c_i\mid q,a)$ are observable. Our pipeline therefore
relies on the controller as a learned proxy for the
training-time selection rule, and operates in three stages.

\smallskip
\textbf{Query-side pre-filtering.}
We first reduce the search space using the precomputed query saliency:
\begin{equation}
\tilde{\mathcal{H}} \;=\; \mathrm{TopK}_{\mathrm{Sal}_q}(\mathcal{H};K),
\end{equation}\label{eq:prefilter} 
$\tilde{\mathcal{H}}$ is presented
to the controller in temporal order, annotated with each turn's
original \texttt{message\_idx}.

\smallskip
\textbf{Controller as learned gate.}
Given $(q,\tilde{\mathcal{H}})$, the controller emits a structured
output consisting of a $\langle\mathrm{thought}\rangle$ block followed by
an $\langle\mathrm{action}\rangle$ JSON object:
\begin{equation}
(\hat g_\tau,\;\hat S_\tau) \;=\; \mathrm{Controller}(q,\tilde{\mathcal{H}}),\qquad
\hat S_\tau \subseteq \tilde{\mathcal{H}},\;\;|\hat S_\tau|\le k,
\end{equation}
where $\hat g_\tau\!\in\!\{0,1\}$ approximates the training gate
$g_\tau\!=\!\mathbf{1}[\,\exists\,c_i \text{ s.t. } y_i\!=\!1\,]$ and
$\hat S_\tau$ approximates the focus set
$F^\star\!=\!\{c_i:y_i\!=\!1\}$.
Crucially, no $\mathrm{Sal}_a$ computation is performed at inference;
the controller has internalized the $\tau$-thresholded selection rule
through supervised training, realizing
\textbf{Bayes-factor-guided memory selection} purely from
self-contained dialogue features.

\smallskip
\textbf{Gated reading.}
The reader is invoked conditionally on the predicted gate:
\begin{equation}
\hat y \;=\;
\begin{cases}
\mathrm{Reader}\!\big(q,\,\mathcal{O},\,\mathcal{H}_{\hat S_\tau}\big) & \text{if }\hat g_\tau=1,\\[2pt]
\mathrm{Reader}\!\big(q,\,\mathcal{O},\,\varnothing\big) & \text{if }\hat g_\tau=0,
\end{cases}
\end{equation}
where $\mathcal{O}$ denotes the four MCQ options and
$\mathcal{H}_{\hat S_\tau}$ the user turns indexed by $\hat S_\tau$,
presented in temporal order. The reader produces a free-form rationale
terminating in a $\langle\mathrm{final\_answer}\rangle$ token from which
the predicted option letter is parsed. When $\hat g_\tau{=}0$ the reader
falls back to query-only answering, which prevents spurious memory
injection on questions whose answer is not memory-bound.

\section{Experiments}
\label{sec:experiments}
\subsection{Experimental Setup}

\paragraph{Datasets.}\label{sec:dataset}
We evaluate on four benchmarks: MemBench-Low/High \citep{tan2025membenchcomprehensiveevaluationmemory},
where Low tests explicit recall and High requires implicit inference; PersonaMem
\citep{jiang2025knowmerespondme}, evaluating personalized responses under evolving user profiles; and PrefEval
\citep{zhao2025llmsrecognizepreferencesevaluating}, evaluating
user-preference recall. We apply a seed-42 stratified 80/10/10 split.
For MB-Low we
report accuracy on the user-only subset ($n{=}374$), excluding
\texttt{*-assistant} scenarios whose evidence falls outside
$\mathcal{H}$. Full details are in Appx.~\ref{app:data}.

\paragraph{Baselines.}
We compare against five retrieval baselines. \textbf{Full context} concatenates all user turns and passes them directly to the reader. \textbf{A-MEM}~\cite{xu2025amemagenticmemoryllm}, \textbf{MemoryBank}~\cite{zhong2023memorybankenhancinglargelanguage}, and \textbf{Mem0}~\cite{chhikara2025mem0buildingproductionreadyai} represent memory-augmented retrieval systems. \textbf{RMM}~\cite{tan-etal-2025-prospect} organizes dialogue history into reflective multi-granularity memory summaries and retrieves evidence via embedding similarity. All retrieved memories are substituted into the same reader prompt as our method. Following prior work, we omit RMM’s retrospective reranking stage and MemoryBank’s forgetting update, as neither is applicable under single-shot MCQ evaluation. For each selector, we evaluate two reader models: \texttt{Qwen3.5-9B-thinking} \citep{qwen35blog} and \texttt{Llama-3.1-8B-Instruct} \citep{grattafiori2024llama3herdmodels}. The $Sal_a$ and $Sal_q$ scores are computed using \texttt{Qwen3.5-0.8B}.

\paragraph{Evaluation Protocol.} We report MCQ accuracy. The reader emits
free-form text terminated by
$\langle\mathrm{final\_answer}\rangle(\ell)$ with
$\ell\!\in\!\{a,b,c,d\}$; we extract the letter and compare to the
gold label. If parsing fails, we fall back to scanning the response
for any unique \texttt{(a)}/\texttt{(b)}/\texttt{(c)}/\texttt{(d)}
mention;
are reported.
MB-Low accuracy uses the user-only subset
($n{=}374$ test records); the other three
benchmarks use all kept test records.

\subsection{Main Results}

\begin{table}[t]
\centering
\small
\begin{tabular}{l lccccc}
\toprule
\textbf{Model} & \textbf{Method} & \textbf{MB-Low} & \textbf{MB-High} & \textbf{Persona} & \textbf{PrefEval} & \textbf{Avg} \\
\midrule
Qwen3.5-9B-Thinking &full ctx&64.17&56.70&36.49&98.99&64.09\\
 & RMM & 36.36 & 54.12 & 50.00 & 79.80 & 57.60 \\
  & Mem0 & 42.51 & 61.34 & 41.89 & 83.84 & 62.86 \\
& A-mem & 53.29 & 64.89 & 55.71 & 86.46 & 65.09\\
& MemoryBank & 51.47 & 69.58 & 52.86 & 88.54 & 65.64\\
 & \textbf{Ours} & \textbf{88.78} & \textbf{77.32} & \textbf{75.68} & \underline{97.98} & \textbf{84.94} \\
\midrule
Llama-3.1-8B-Instruct &full ctx& 55.35  & 56.19 & 48.65 & 78.79 & 59.74\\
 & RMM & 24.60 & 61.34 & 40.54 & 88.89 & 53.84 \\
 & Mem0 & 38.88 & 61.34 & 58.11 & 94.95 & 63.32 \\
& A-mem &56.95& 69.07&66.22& 92.93& 71.29 \\
& MemoryBank & 57.49 & 71.65 & 67.57 & 89.90 & 71.65\\
 & \textbf{Ours} & \textbf{82.62} & 68.56 & \textbf{77.03} & \textbf{99.99} & \textbf{82.05} \\
\bottomrule
\end{tabular}
\caption{Main results and ablations. Best results per benchmark are in \textbf{bold}.}
\label{tab:main}
\end{table}

\paragraph{Our Method Achieves the Best Overall Performance.} Tab.~\ref{tab:main} summarizes performance across four benchmarks. Across both reader backbones, our method consistently achieves the strongest average accuracy, substantially outperforming all retrieval-augmented baselines as well as the full-context setting. Under Qwen3.5-9B-Thinking, our method achieves an average score of 84.94, surpassing the strongest baseline, MemoryBank (65.64), by +19.30 points, while improving over the full-context baseline (64.09) by +20.85 points. Similar trends hold under Llama-3.1-8B-Instruct, where our method reaches 82.05 average accuracy compared to 71.65 for MemoryBank and 59.74 for full-context prompting.

\paragraph{Gains are concentrated on inference-heavy benchmarks.}
Performance gains are especially pronounced on MB-High and Persona, where our method consistently achieves the strongest improvements across both reader backbones. Under Qwen3.5-9B-Thinking, our method improves from 69.58 to 77.32 on MB-High (+7.74) and from 55.71 to 75.68 on Persona (+19.97) relative to the strongest retrieval baseline, MemoryBank. Similar trends hold under Llama-3.1-8B-Instruct, where our method improves from 67.57 to 77.03 on Persona (+9.46). These benchmarks share a common structure: the answer depends on recovering implicit, temporally distant, or weakly lexicalized preference evidence rather than retrieving a single semantically aligned turn. 
By contrast, the gap narrows on PrefEval, where preferences are often stated explicitly and semantic retrieval is already effective. Under Qwen3.5-9B-Thinking, MemoryBank reaches 88.54 while our method achieves 97.98 (+9.44). These results suggest that the advantage of our method is largest when retrieval depends on identifying high-utility evidence among semantically related but uninformative history.

Notably, our method also consistently outperforms the full-context baseline despite conditioning on substantially fewer dialogue turns. Under Qwen3.5-9B-Thinking, full-context prompting achieves only 64.09 average accuracy compared to our 84.94 (+20.85), while under Llama-3.1-8B-Instruct the gap is similarly large (59.74 vs.\ 82.05, +22.31). This suggests that the gains do not arise from increased memory access alone, but from selectively isolating the small subset of interactions that genuinely contribute to the downstream response. Together, these results support the view that effective personalized memory retrieval depends less on maximizing topical coverage than on identifying evidence with high answer utility.

\begin{figure}[t]
    \centering
    \includegraphics[width=\columnwidth]{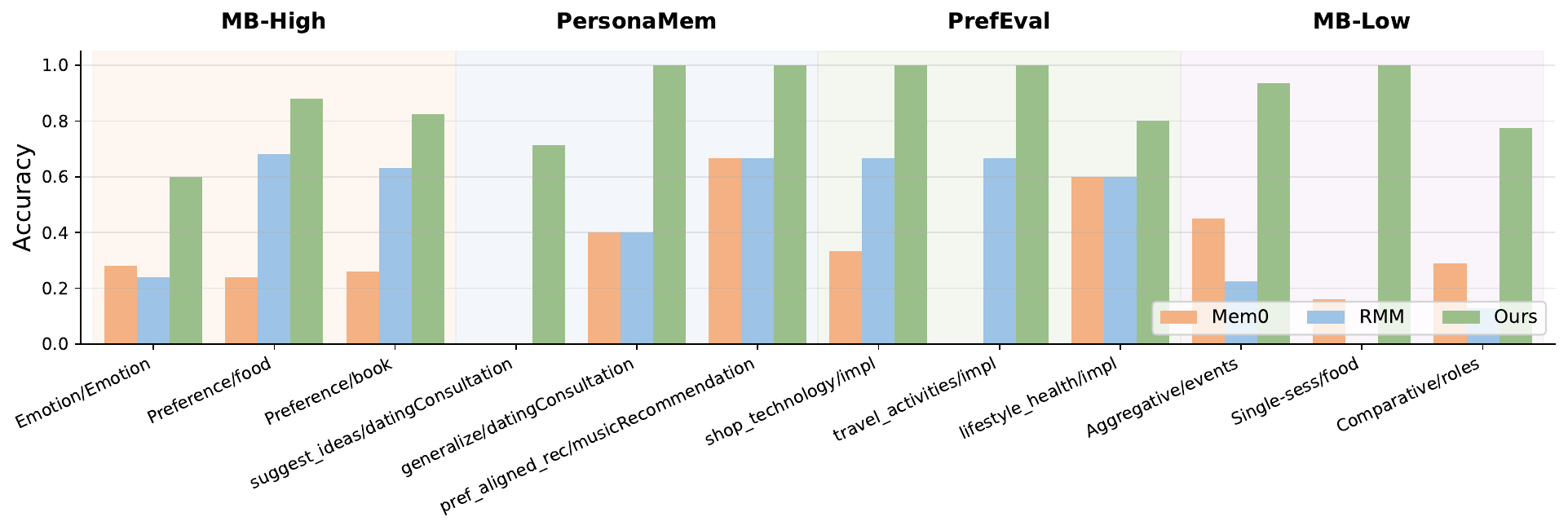}
    \caption{\textbf{Sub-category breakdown across four benchmarks.} Accuracy of Mem0, RMM, and our method on twelve representative sub-categories;three per benchmark, drawn from MB-High, PersonaMem, PrefEval, and MB-Low. Sub-categories are selected to span the regimes where preference evidence is implicit (PersonaMem dating/music consultation, PrefEval implicit-choice), aggregated across scattered turns (MB-Low \texttt{Aggregative/events}, \texttt{Comparative/roles}), or buried in long context (MB-High \texttt{Preference/\{book,food\}}, \texttt{Emotion}). Our method outperforms the stronger baseline by 20 to 77 absolute points on every selected sub-category.}

    \label{fig:group}
\end{figure}

Fig.~\ref{fig:group} compares our method against Mem0 and RMM across twelve representative sub-categories. The largest gains appear on PersonaMem dating/recommendation tasks (+33 to +71) and MB-Low aggregative/comparative tasks (+48 to +77), where answers depend on sparse or distributed evidence rather than direct semantic overlap. MB-High and PrefEval further highlight the difficulty of long-context preference and sentiment recall, where relevant evidence is often overwhelmed by topically related but evidentially irrelevant dialogue.  Conversely, when retrieval reduces to straightforward query--passage similarity, RMM is already competitive and the remaining improvement becomes much smaller. These patterns support our central claim that effective personalized memory retrieval depends less on topical overlap than on identifying evidence with genuine answer utility.

\begin{figure}[t]
    \centering
\includegraphics[width=\columnwidth]{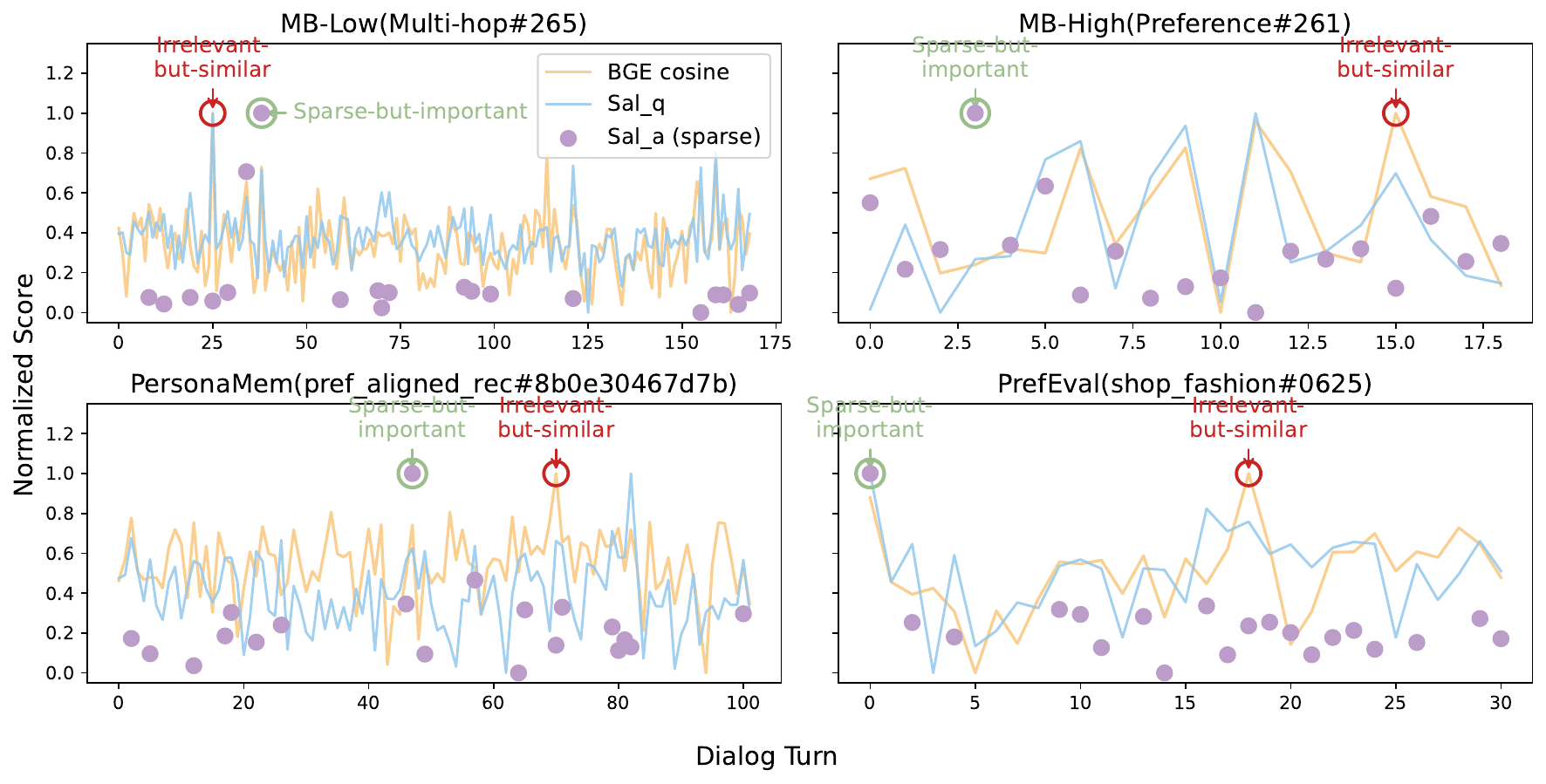}
    \caption{Per-benchmark comparison of retrieval signals over a single conversation.
    BGE peaks \textcolor[HTML]{C82423}{Irrelevant-but-similar} tend to land on turns that are lexically aligned with the query but carry no evidence for the gold answer, whereas $Sal_a$ peaks \textcolor[HTML]{9BBF8A}{Sparse-but-important} concentrate on the few turns that actually shift the best answer likelihood.
    }
    \label{fig:sal_viz}
\end{figure}

\textbf{Bayes factors capture evidential rather than topical relevance.}
Fig.~\ref{fig:sal_viz} visualizes the retrieval distributions induced by semantic similarity and saliency-based scoring across four representative benchmarks, with additional qualitative retrieval examples provided in Appx. Tab.~\ref{tab:bge-vs-sala}. Across all cases, BGE assigns high scores to turns that are lexically or topically aligned with the query, even when those turns contribute little evidence toward resolving the underlying preference or decision. In contrast, $Sal_a$ consistently concentrates on a small number of temporally sparse yet decision-critical turns, often prioritizing evidence that semantic retrieval ranks far below the top. This contrast reflects the different retrieval objectives induced by the two signals: semantic similarity favors surface alignment, whereas Bayes-factor saliency measures whether conditioning on a memory turn materially shifts the model’s belief toward the correct response. Effective personalized memory retrieval therefore depends less on topical relatedness than on identifying the sparse pieces of context that carry genuine evidential utility.

\subsection{Ablation Study}\label{sec:ab}

\begin{table}[htbp]
\centering
\setlength{\tabcolsep}{4pt}
\begin{tabular}{p{3.2cm}rrrrr}
\toprule
\textbf{Method} & \textbf{MB-Low} & \textbf{MB-High} & \textbf{Persona} & \textbf{PrefEval} & \textbf{Avg} \\
\midrule

Qwen3.5 + PPL    
& 25.67 {\scriptsize(-63.11)} 
& 57.22 {\scriptsize(-20.10)} 
& 39.19 {\scriptsize(-36.49)} 
& 40.40 {\scriptsize(-57.58)} 
& 40.62 {\scriptsize(-44.32)} \\

\hspace{3.65em} + BGE     
& 59.63 {\scriptsize(-29.15)} 
& 59.28 {\scriptsize(-18.04)} 
& 56.76 {\scriptsize(-18.92)} 
& 73.74 {\scriptsize(-24.24)} 
& 62.35 {\scriptsize(-22.59)} \\

\hspace{3.65em} + full ctx 
& 64.17 {\scriptsize(-24.61)} 
& 56.70 {\scriptsize(-20.62)} 
& 36.49 {\scriptsize(-39.19)}
& \textbf{98.99} {\scriptsize(+1.01)} 
& 64.09 {\scriptsize(-20.85)} \\

\hspace{3.65em} + $Sal_a$  
& 60.16 {\scriptsize(-28.62)} 
& 64.95 {\scriptsize(-12.37)} 
& 59.46 {\scriptsize(-16.22)} 
& 79.80 {\scriptsize(-18.18)} 
& 66.09 {\scriptsize(-18.85)} \\

\textbf{Ours}         
& \textbf{88.78} 
& \textbf{77.32} 
& \textbf{75.68} 
& 97.98 
& \textbf{84.94} \\

\bottomrule
\end{tabular}
\caption{Main results on Qwen3.5-9B-think.
Methods differ only in how evidence is selected from the
user-turn history: \textit{PPL} ranks turns by per-turn
cross-entropy, \textit{BGE} by query--turn cosine similarity,
\textit{full ctx} uses every user turn, and
\textit{$Sal_{a}$} is the oracle answer-conditioned
saliency upper bound. Numbers are accuracy (\%); parentheses indicate
the gap to \textbf{Ours} in percentage points.}
\label{tab:ab1}
\end{table}

\begin{table}[t]
\centering
\setlength{\tabcolsep}{4pt}
\begin{tabular}{lccccc}
\toprule
\textbf{Method} & \textbf{MB-Low} & \textbf{MB-High} & \textbf{Persona} & \textbf{PrefEval} & \textbf{Avg} \\
\midrule

EMB labels ($\text{replace }$ $Sal_a$) 
& 89.98 {\scriptsize(+1.20)} 
& 68.56 {\scriptsize(-8.76)} 
& 75.68 {\scriptsize(0.00)} 
& 95.96 {\scriptsize(-2.02)} 
& 82.55 {\scriptsize(-2.39)} \\

w/o Controller 
& 89.58 {\scriptsize(+0.80)} 
& 73.71 {\scriptsize(-3.61)} 
& 68.92 {\scriptsize(-6.76)} 
& \textbf{99.99} {\scriptsize(+2.01)} 
& 84.90 {\scriptsize(-0.04)} \\

\textbf{Ours} 
& 88.78 
& \textbf{77.32} 
& \textbf{75.68} 
& 97.98 
& \textbf{84.94} \\

\bottomrule
\end{tabular}
\caption{Ablations on Qwen3.5-9B-think.
During the supervised fine-tuning stage,
\textit{EMB labels} replaces the $Sal_{a}$ silver labels
with query--turn embedding similarity, isolating the supervision
signal. \textit{w/o Controller} removes the controller and selects
evidence directly from the saliency stage, isolating the
controller during training. Numbers are accuracy (\%); parentheses indicate
the gap to \textbf{Ours} in percentage points.}
\label{tab:ab2}
\end{table}

We conduct three ablations to assess the contribution of each component.

\textbf{Supervised fine-tuning is essential}.
According to Tab.~\ref{tab:ab1},
removing supervised fine-tuning and using the base Qwen3.5-9B-Thinking model with $\mathrm{Sal}_a$ scoring at inference time collapses performance to 66.09 avg.
This demonstrates that the utility signal must be internalized through training rather than applied post-hoc; an untuned model cannot reliably identify high-utility memory turns at inference.

\textbf{$Sal_a$ is the critical signal} (Tab.~\ref{tab:ab2}).
Replacing $Sal_a$-derived labels with embedding-based pseudo-labels (using cosine similarity to identify relevant turns) reduces average accuracy from 84.94 to 82.55 ($-2.39$ points). The gap is most pronounced on MB-High (77.32 $\rightarrow$ 68.56, $-8.76$ points), the benchmark demanding the most implicit preference inference. This confirms that $Sal_a$ captures a qualitatively different supervision signal from semantic similarity: it reflects whether a memory turn actually reduces answer perplexity, rather than whether it is topically close to the query.

\textbf{The controller provides marginal but consistent benefit} (Tab.~\ref{tab:ab2}).
Removing the controller while retaining other components yields 84.90 avg, a negligible drop of 0.04 points overall. However, per-benchmark analysis reveals that the controller provides a more consistent benefit on MB-High (77.32 vs.\ 73.71, $+3.61$ points), suggesting its value is concentrated in harder retrieval scenarios where gating decisions are more consequential.

\textbf{Retrieval vs.\ reader adaptation.} The performance gap between our method and retrieval-only
baselines should not be interpreted as a retrieval gain alone.
Empirically, the oracle selector $Sal_a$ combined with the
thinking-enabled reader recovers much of the performance,
suggesting that our learned
$\mathrm{Sal}_q \!\rightarrow$ controller $\!\rightarrow$ focus-set
pipeline approximates the oracle retrieval distribution.
This indicates that much of the remaining improvement
comes from jointly training the reader and retrieval
controller, rather than retrieval quality alone.

We view this as an important property of the framework rather than a
confounder. Existing memory systems such as Mem0 or RMM rely on
fixed extraction pipelines, external memory stores, or online
reflection policies that are not naturally amenable to end-to-end
supervised fine-tuning. In contrast, our framework jointly trains retrieval and reading,
allowing the reader to adapt to compressed evidence representations.
Thus, the observed gains arise not only from better evidence
selection, but also from co-adaptation between memory selection and
downstream reasoning.

\section{Conclusion}
We proposed a Bayes-factor-based framework for personalized memory retrieval that unifies memory access gating and turn selection under a single principled signal. Unlike embedding-based approaches that measure topical similarity, our method estimates the evidential utility of each memory turn for predicting the correct answer. By jointly training a controller and reader on Bayes-factor-derived proxy labels, the framework learns when to retrieve and what to retrieve without requiring gold answers at inference time. Experiments on four personalization benchmarks show consistent improvements over existing memory systems, with the largest gains on long-context, implicit-preference tasks where semantic similarity fundamentally fails. These results highlight the importance of optimizing for answer utility rather than topical relatedness in memory-augmented dialogue systems.

\newpage
\bibliographystyle{unsrtnat}
\bibliography{ref}


\appendix

\section{Dataset preprocessing}\label{app:data}

\paragraph{Filtering.}
After running the saliency stage of
Section~\ref{sec:sala}--\ref{sec:threshold}, instances for which
$|S_\tau|=0$ \emph{and} the maximum reach-top-1 log-probability
falls below a secondary threshold are dropped, as these contain no
usable evidence even at the turn level.

\paragraph{Splits.}
Remaining instances are partitioned with a fixed seed-42 stratified
$80/10/10$ split. PersonaMem additionally enforces persona
disjointness: all $(q,\mathcal{H})$ pairs sharing the same
\texttt{persona\_id} stay in the same split, preventing test-time
leakage of memorized persona traits across the 20 personas
(14 train / 3 val / 3 test). Per-benchmark counts are listed in
Table~\ref{tab:splits}.

\paragraph{MB-Low \texttt{*-assistant} scenarios.}
$25\%$ of MB-Low test records (124/499) lie in
\texttt{Single-session-assistant} or \texttt{Multi-session-assistant}
scenarios, which ask about the assistant's past
\emph{recommendations} rather than the user's own statements.
Because $\mathcal{H}$ in our setup contains only user turns, the
answer content for these records lies in assistant turns that have
been removed from the evidence pool, making them ill-posed under
our problem formulation (Section~\ref{sec:setup}). We therefore
report MB-Low accuracy on the user-only subset ($n{=}374$); a
full-context formulation that retains assistant turns would be
required for fair evaluation on \texttt{*-assistant} scenarios and
is beyond this work's scope.

\begin{table}[h]
\centering\small
\setlength{\tabcolsep}{4pt}
\begin{tabular}{lrrrr}
\toprule
& \textbf{MB-Low} & \textbf{MB-High} & \textbf{PersonaMem} & \textbf{PrefEval} \\
\midrule
Train (kept)  & 1{,}871 & 977 & 431 & 801 \\
Val (kept)    & 375     & 194 & 84  & 100 \\
Test (kept)   & 374     & 194 & 74  & 99  \\
\midrule
$|\mathcal{D}_{\text{ctrl}}\!\cup\!\mathcal{D}_{\text{read}}|$ (Train) & 4{,}069 & 2{,}358 & 1{,}022 & 1{,}932 \\
\quad of which $g_\tau\!=\!1$ & 1{,}628 & 943 & 409 & 773 \\
\quad of which $g_\tau\!=\!0$ & 813 & 472 & 204 & 386 \\
\bottomrule
\end{tabular}
\caption{Per-benchmark split sizes after the saliency drop filter.
For MB-Low we additionally exclude \texttt{*-assistant} scenarios
(25\% of records) at the dataset construction stage, so that the
training distribution matches the user-only evaluation protocol
(\S\ref{sec:dataset}). Each kept training instance contributes one
controller sample and, when $g_\tau\!=\!1$, one reader sample.
Negative controller samples ($g_\tau\!=\!0$) are described in
\S\ref{sec:train_pipeline}.}
\label{tab:splits}
\end{table}

\section{Implementation Details}\label{app:implement}

\paragraph{Base model and fine-tuning.}
Our pipeline fine-tunes \textbf{Qwen3.5-9B} jointly on
$\mathcal{D}_{\text{ctrl}}\!\cup\!\mathcal{D}_{\text{read}}$ across the
four benchmarks (full-parameter, no LoRA). We use DeepSpeed ZeRO-3 \citep{rajbhandari2020zeromemoryoptimizationstraining}
across 4 NVIDIA Ada A6000 GPUs, learning rate $5\!\times\!10^{-6}$, cosine schedule,
warm-up ratio $0.03$, $1$ epoch, effective batch size $16$
(\texttt{per\_device\_batch\_size}$=1$, gradient accumulation $2$),
cutoff length $2{,}048$, bfloat16, and gradient checkpointing.
Training runs for 1{,}343 optimizer steps and is conducted with the
\texttt{qwen3\_nothink} chat template; the only thinking-enabled
runs are inference-time baselines.

\paragraph{Saliency stage.}
$\mathrm{Sal}_a$ and $\mathrm{Sal}_q$ are computed offline with the
same Qwen3.5-0.8B model; per-benchmark thresholds $\tau$ are
selected on the validation split such that $|S_\tau|>0$ for
$\geq\!95\%$ of records and the median $|S_\tau|$ stays in the
$5$--$15$ range across benchmarks. Within each
$S_\tau$, the reader's focus set is capped at top-$k$ with $k{=}5$
(\texttt{focus\_chunk\_indices}, Section~\ref{sec:learning}).

\paragraph{Negative controller sample}.
To prevent the controller from collapsing, the training set is balanced with
\emph{negative controller samples} where $S_\tau\!=\!\varnothing$ and
$g_\tau\!=\!0$. These arise naturally from instances whose Sal\textsubscript{a}
distribution is uniformly below $\tau$, and are augmented by
cross-instance mismatches (the query of one instance paired with the
history of another, ensuring no turn carries answer-relevant evidence).

\paragraph{Inference compute.}
Pre-filtering takes $K{=}20$ candidates by $\mathrm{Sal}_q$ for
MB-Low, MB-High, and PrefEval, and $K{=}10$ for PersonaMem
(PersonaMem turns are substantially longer per turn, so a smaller
$K$ keeps the controller prompt within context length).
All inference uses Hugging Face \texttt{transformers} with greedy
decoding (\texttt{do\_sample=False}) and \texttt{max\_new\_tokens=120}
for direct MCQ answering; thinking-enabled variants extend this to
$2{,}048$ to accommodate the $\langle\text{think}\rangle\dots$ trace.

\section{Retrieval Behavior Case Study}
\begin{table*}[h]
\centering
\small
\setlength{\tabcolsep}{4pt}
\renewcommand{\arraystretch}{1.25}
\begin{tabular}{p{0.10\linewidth} p{0.20\linewidth} p{0.30\linewidth} p{0.30\linewidth}}
\toprule
\textbf{Benchmark} & \textbf{User query} & \textbf{BGE \textcolor[HTML]{C82423}{(irrelevant-but-similar)}} & \textbf{Sal$_a$ \textcolor[HTML]{9BBF8A}{(sparse-but-important)}} \\
\midrule
MB-Low &
\textit{What is the company name for someone whose hobby is fitness?} &
\textit{``My cousin is really into fitness; he loves working out and staying healthy.''}\newline
\textcolor[HTML]{C82423}{\footnotesize Lexically matches ``fitness'', but contains no company name.}
&
\textit{``My cousin's name is Emerald City Grocers, and he's doing some great work with his company.''}\newline
\textcolor[HTML]{9BBF8A}{\footnotesize Carries the actual answer (the cousin's company name).}
\\
\midrule
MB-High &
\textit{According to the movies I mentioned, what kind of movies might I prefer to watch?} &
\textit{``I'm really into `The Empire Strikes Back,' but I also love `A Room with a View'\dots''}\newline
\textcolor[HTML]{C82423}{\footnotesize Mentions specific movie titles; topically aligned but does not state a preference type.}
&
\textit{``And the romance is just perfect! It's all about fleeting moments and bittersweet encounters.''}\newline
\textcolor[HTML]{9BBF8A}{\footnotesize Reveals the latent preference dimension (\emph{romance}) the question asks about.}
\\
\midrule
PersonaMem &
\textit{I'm planning a weekend getaway and want to try something creatively fulfilling. What would you suggest?} &
\textit{``Recently, I visited a singles club that organized some fun activities\dots''}\newline
\textcolor[HTML]{C82423}{\footnotesize Surface match on ``fun activities''; says nothing about user's creative direction.}
&
\textit{``Blend my love for the outdoors with my musical creativity\,---\,the sounds of nature really added depth to the track.''}\newline
\textcolor[HTML]{9BBF8A}{\footnotesize Captures the recent shift toward outdoor / nature recording that justifies the gold recommendation.}
\\
\midrule
PrefEval &
\textit{What kind of jacket should I buy for the upcoming winter?} &
\textit{``Got any philosophical advice?''}\newline
\textcolor[HTML]{C82423}{\footnotesize No lexical or semantic relation; BGE picks it from spurious embedding noise.}
&
\textit{``I exclusively wear cruelty-free, vegan fashion made from sustainable materials like organic cotton, bamboo, and recycled plastics.''}\newline
\textcolor[HTML]{9BBF8A}{\footnotesize States the user's hard preference constraint that the recommended jacket must satisfy.}
\\
\bottomrule
\end{tabular}
\caption{For each benchmark, the top-ranked turn by semantic embedding (BGE) versus by answer-conditioned saliency ($\mathrm{Sal}_a$). Across all four cases, BGE peaks on turns that are lexically or topically aligned with the query but carry no evidence for the gold answer (\emph{irrelevant-but-similar}), while $\mathrm{Sal}_a$ identifies the sparse turns that actually justify the correct response (\emph{sparse-but-important}).}
\label{tab:bge-vs-sala}
\end{table*}

\section{Proofs of Theoretical Results}
\paragraph{The realistic case: bounded approximation.}\label{app:proofs}
Strict query-conditional sufficiency rarely holds. We give a quantitative bound that controls the disagreement between $\mathrm{Sal}_q$- and $\mathrm{Sal}_a$-induced rankings in terms of how badly sufficiency is violated.

\begin{theorem}[Bounded ranking disagreement]
\label{thm:bound}
Define the sufficiency residual of turn $c_i$ as
\begin{equation}
\varepsilon_i \;:=\; \bigl| \log p(a \mid q, c_i) - \log p(a \mid c_i) \bigr|.
\end{equation}
Then for any pair $c_i, c_j$,
\begin{equation}
\Bigl| \bigl[\mathrm{Sal}_a(c_i) - \mathrm{Sal}_a(c_j)\bigr] - \bigl[\mathrm{Sal}_q(c_i) - \mathrm{Sal}_q(c_j)\bigr] \Bigr|
\;\le\;
\varepsilon_i + \varepsilon_j + \bigl| \mathrm{Sal}_a^{\mathrm{marg}}(c_i) - \mathrm{Sal}_a^{\mathrm{marg}}(c_j) \bigr|.
\label{eq:bound}
\end{equation}
\end{theorem}

\noindent Theorem~\ref{thm:bound} pins down two distinct sources of surrogate error. The first, $\varepsilon_i + \varepsilon_j$, measures how much additional information the query carries about the answer \emph{beyond} what $c_i$ already provides; this term is small for declarative preference turns and large when the gold answer requires multi-turn aggregation that no single $c_i$ resolves.The second, the marginal-evidence gap, is suppressed by the query-side pre-filter.
We show them empirically in Fig.~\ref{fig:sal_viz}.

For each validation example $\times$ top-20 $\mathrm{Sal}_q$ chunks (878 examples, 16{,}275 chunks total), we computed $|\log p(a \mid q, c_i) - \log p(a \mid c_i)|$ using Qwen3.5-9B with the saliency-pipeline prompt format, comparing the chunk-only conditional against the chunk-plus-question conditional. Results show the surrogate gap is highly benchmark-dependent: \texttt{persona} is near-faithful (mean $0.47$ nats; $98.7\%$ of chunks within $1$ nat), \texttt{mbhl} moderate (mean $1.67$, $p_{90} = 3.42$), \texttt{prefeval} has a tight unimodal gap (mean $1.32$, $p_{90} = 2.04$), and \texttt{mbll} is heavy-tailed (mean $2.30$, $p_{90} = 6.80$, with a clear bimodal shape). This confirms that the $q \perp a \mid c_i$ assumption holds where chunks carry declarative content, but $\varepsilon_i$ is non-trivial precisely on the implicit / multi-turn-inference cases (\texttt{mbhl}, \texttt{mbll} right tail) --- which we re-frame the SFT controller as empirically correcting.

\begin{figure}[t]
  \centering
  \includegraphics[width=0.95\linewidth]{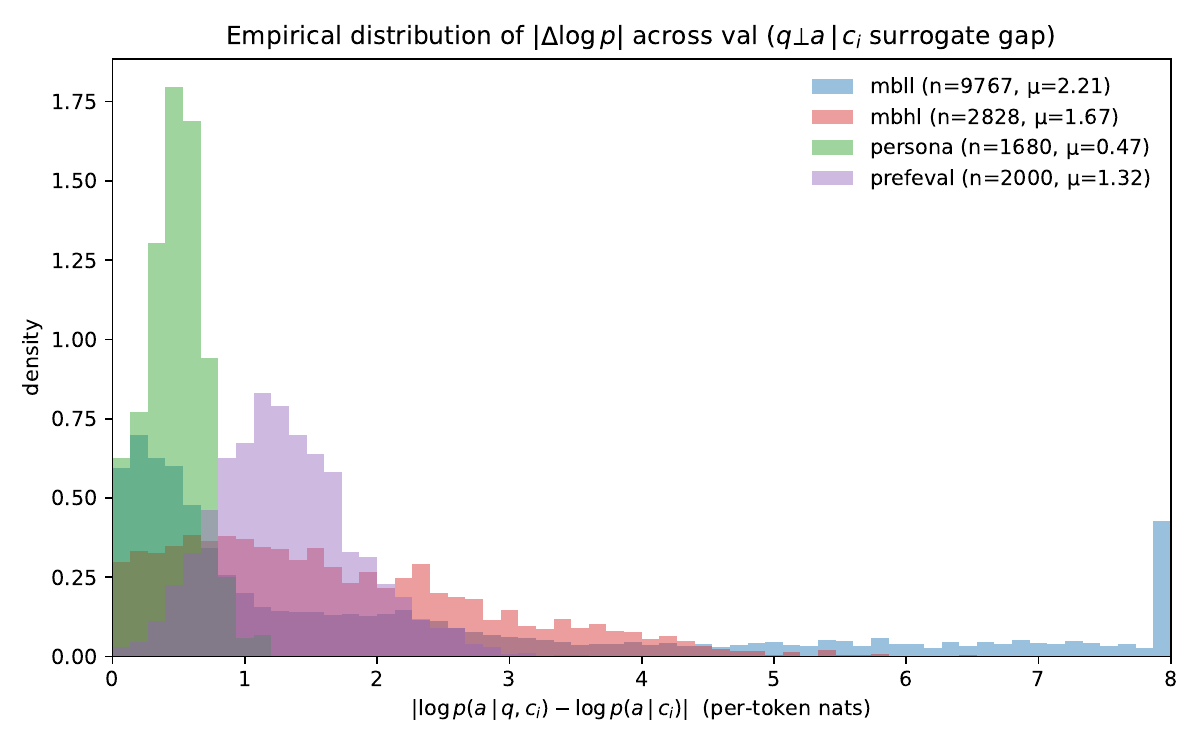}
  \caption{Empirical distribution of the query-conditional surrogate gap $|\log p(a \mid q, c_i) - \log p(a \mid c_i)|$ on the validation set, evaluated with Qwen3.5-9B over the top-20 $\mathrm{Sal}_q$ chunks per example. \texttt{persona} is near-faithful to the $q \perp a \mid c_i$ assumption; \texttt{mbhl} and the right tail of \texttt{mbll} expose non-trivial $\varepsilon_i$, where the SFT controller is empirically rescuing the surrogate.}
  \label{fig:q-indep-hist-combined}
\end{figure}

\begin{figure}[t]
  \centering
  \includegraphics[width=0.98\linewidth]{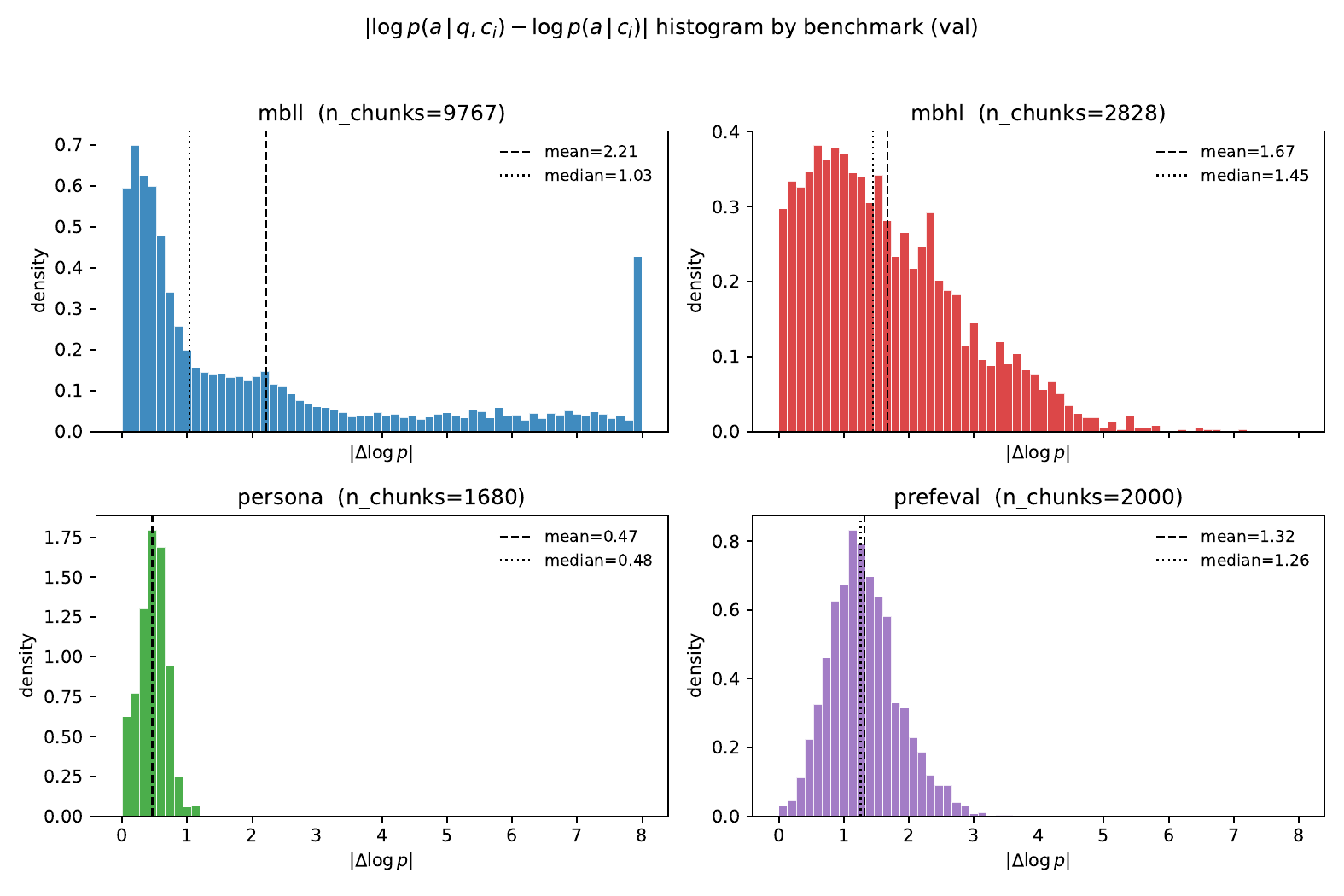}
  \caption{Per-benchmark histograms of $|\log p(a \mid q, c_i) - \log p(a \mid c_i)|$. Dashed/dotted lines mark the mean and median; \texttt{mbll} clips at $|\Delta\log p|=8$ to keep the heavy right tail visible.}
  \label{fig:q-indep-hist-per-bench}
\end{figure}

\definecolor{boxbg}{RGB}{245,245,250}

\newtcolorbox{sftbox}[1]{
  colback=boxbg,
  colframe=black!50,
  fonttitle=\bfseries,
  title=#1,
  breakable,
  left=2pt,right=2pt,top=2pt,bottom=2pt,
  before skip=4pt,after skip=4pt
}

\lstdefinestyle{sft}{
  basicstyle=\ttfamily\footnotesize,
  breaklines=true,
  breakatwhitespace=false,
  columns=fullflexible,
  keepspaces=true,
  showstringspaces=false,
  frame=none,
  xleftmargin=0pt,
}

\section{Limitaton}
Although the proposed framework demonstrates strong performance on long-context personalization tasks, it still has several limitations. First, the saliency signals $Sal_a$ and $Sal_q$ are computed with a single small scoring model (Qwen3.5-0.8B). Second, our theoretical guarantees for the proxy signal $Sal_q$ hold most tightly under query-conditional sufficiency (Theorem 3), and instances requiring substantial cross-turn evidence aggregation lie outside this regime; the trained controller mitigates but does not eliminate this gap. In addition, the framework depends on supervised fine-tuning with a jointly trained controller and reader, which adds training complexity and computational cost. Finally, the current formulation only models user-side memories and excludes assistant turns, limiting applicability in fully interactive conversational environments.

\section{SFT Training Data Examples}
\label{app:sft-examples}
\paragraph{Common system / instruction prompts.}
All controller samples share one system prompt; all MCQ samples share another:

\begin{lstlisting}[style=sft]
[MCQ system]
You solve personalization multiple-choice questions. Use the salient
evidence from dialogue history to choose the best option and end with
<final_answer>(a/b/c/d).

[MCQ instruction]
Use the question, answer options, and salient dialogue evidence to
choose the best option. End with <final_answer>(a), (b), (c), or (d).

[Controller system]
You are a personalization controller for a dialogue model. Given the
current question and candidate evidence, decide whether the
memory-reasoning module should use the evidence.

[Controller instruction]
Based on the user's query and the following evidence, which evidence
should be updated to the memory for a best answer quality?
\end{lstlisting}

\lstdefinestyle{sft}{
  basicstyle=\ttfamily\footnotesize,
  breaklines=true,
  breakatwhitespace=false,
  columns=fullflexible,
  keepspaces=true,
  showstringspaces=false,
  frame=none,
  xleftmargin=0pt,
}

\begin{sftbox}{MBLL -- Controller sample (\texttt{qid=Single-hop-roles-353})}

\textbf{Input}
\begin{lstlisting}[style=sft]
current_user_query:
What is the name of my subordinate's company?

dialog_history (time-sorted):

[1] message_idx=19
user: I'm really proud of her achievements. She works so hard.

[2] message_idx=21
user: I want to discuss my subordinate, Oliver Prescott. He's 22 years old.

[3] message_idx=23
user: True! Going back to his age, he has a birthday coming up soon, on May 8th.

[4] message_idx=30
user: Exactly! His contact number is 71806262336, but I mostly communicate with him through email.

[5] message_idx=34
user: My subordinate works at a place called Innovative Tech Engineering Solutions.

[6] message_idx=39
user: I think I should set up a one-on-one meeting to discuss his career goals.

[7] message_idx=43
user: My sister works in Portland, OR.

[8] message_idx=60
user: I'd like to discuss my boss, Juliet Harrington. She's quite impressive.

[9] message_idx=67
user: Yes, I was thinking of organizing a small celebration at the office.

[10] message_idx=74
user: My boss is from Denver, Colorado.

[11] message_idx=76
user: And I also have her email: juliet.harrington@innovativetecheng.com.

[12] message_idx=77
user: I enjoy working under her direction.

[13] message_idx=79
user: I'm looking forward to seeing how we can grow as a team.

[14] message_idx=80
user: Definitely. I feel optimistic about the future.

[15] message_idx=104
user: Her birthday is on June 23rd.

[16] message_idx=119
user: My niece's contact number is 617-091-08991.

[17] message_idx=126
user: I'm thinking of organizing a small surprise party.

[18] message_idx=129
user: My coworker is Mira Caldwell.

[19] message_idx=131
user: Mira works as a Junior Mechanical Engineer at Innovative Tech Engineering Solutions.

[20] message_idx=155
user: I agree! He's quite insightful.
\end{lstlisting}

\textbf{Output}
\begin{lstlisting}[style=sft]
<thought>
The dialog history contains turns relevant to the user's query. Trigger memory-guided reasoning and focus on the listed turns.
</thought>
<action>
{
  "trigger": true,
  "focus_chunk_indices": [
    34,
    131,
    76,
    80,
    19
  ],
  "policy": "use_relevant_chunks_for_memory_reasoning"
}
</action>
\end{lstlisting}

\end{sftbox}

\begin{sftbox}{MBLL -- MCQ sample (\texttt{qid=Single-hop-roles-353})}

\textbf{Input}
\begin{lstlisting}[style=sft]
user_question:
What is the name of my subordinate's company?

candidate_options:
(A) Innovative Tech Engineering Solutions
(B) Innovative Engineering Solutions
(C) Tech Engineering Innovations
(D) Creative Tech Solutions

salient_evidence:

[1] message_idx=19
user: I'm really proud of her achievements. She works so hard.

[2] message_idx=34
user: My subordinate works at a place called Innovative Tech Engineering Solutions.

[3] message_idx=76
user: And I also have her email: juliet.harrington@innovativetecheng.com.

[4] message_idx=80
user: Definitely. I feel optimistic about the future of our projects.

[5] message_idx=131
user: Mira works as a Junior Mechanical Engineer at Innovative Tech Engineering Solutions.
\end{lstlisting}

\textbf{Output}
\begin{lstlisting}[style=sft]
Innovative Tech Engineering Solutions

<final_answer>(a)
\end{lstlisting}

\end{sftbox}

\begin{sftbox}{MBHL -- Controller sample (\texttt{qid=Preference-food-394})}

\textbf{Input}
\begin{lstlisting}[style=sft]
current_user_query:
According to the dishes I mentioned, Which flavor I might prefer?

dialog_history (time-sorted):

[1] message_idx=0
user: I'm a big fan of Hot Mulled Wine; there's just something about that warm, spiced flavor that really hits the spot.

[2] message_idx=1
user: I really enjoy those cozy drinks during wintertime gatherings.

[3] message_idx=2
user: There's something about spices and citrus that just works so well together.

[4] message_idx=3
user: And I love how it fills the room with a delightful aroma!

[5] message_idx=4
user: It's perfect for sharing with friends as the cold weather sets in.

[6] message_idx=5
user: I'm a big fan of Hot Mulled Wine, but I also really enjoy a good Spicy Chai Latte.

[7] message_idx=6
user: I've been craving a warm drink that has a perfect blend of spices and creamy texture.

[8] message_idx=7
user: I love how the flavors can be both bold and sweet at the same time, creating a unique experience.

[9] message_idx=8
user: I'm really into Warm Cinnamon Rolls, but I also love Spicy Chai Latte.

[10] message_idx=9
user: There's something magical about that warm, pillowy texture that melts in your mouth.

[11] message_idx=10
user: And I love how the aroma fills the kitchen, inviting everyone to come and enjoy.
\end{lstlisting}

\textbf{Output}
\begin{lstlisting}[style=sft]
<thought>
The dialog history contains turns relevant to the user's query. Trigger memory-guided reasoning and focus on the listed turns.
</thought>
<action>
{
  "trigger": true,
  "focus_chunk_indices": [
    8,
    0,
    5,
    6,
    1
  ],
  "policy": "use_relevant_chunks_for_memory_reasoning"
}
</action>
\end{lstlisting}

\end{sftbox}

\begin{sftbox}{MBHL -- MCQ sample (\texttt{qid=Preference-food-394})}

\textbf{Input}
\begin{lstlisting}[style=sft]
user_question:
According to the dishes I mentioned, Which flavor I might prefer?

candidate_options:
(A) Bitter
(B) Sweet and Spicy
(C) Warm and Spicy
(D) Sweet

salient_evidence:

[1] message_idx=0
user: I'm a big fan of Hot Mulled Wine; there's just something about that warm, spiced flavor that really hits the spot.

[2] message_idx=1
user: I really enjoy those cozy drinks during wintertime gatherings.

[3] message_idx=5
user: I'm a big fan of Hot Mulled Wine, but I also really enjoy a good Spicy Chai Latte.

[4] message_idx=6
user: I've been craving a warm drink that has a perfect blend of spices and creamy texture.

[5] message_idx=8
user: I'm really into Warm Cinnamon Rolls, but I also love Spicy Chai Latte.
\end{lstlisting}

\textbf{Output}
\begin{lstlisting}[style=sft]
Warm and Spicy

<final_answer>(c)
\end{lstlisting}

\end{sftbox}


\begin{sftbox}{Persona -- Controller sample (\texttt{qid=4a1d9313-7e19-446d-b3fa-47f7e7b06b90})}

\textbf{Input}
\begin{lstlisting}[style=sft]
current_user_query:
I'm considering diving into exploring how authors incorporate societal influences into their storytelling. Any suggestions on how I could engage with this creatively?

dialog_history (time-sorted):

[1] message_idx=1
user: The discussions are very enriching. They really allow for a deep dive into the themes and character motivations that the authors weave into their narratives. Each member brings unique perspectives, which often highlight angles of interpretation I hadn't considered before. Such interaction fosters a sense of community and camaraderie among us, and it's wonderful to be part of a group where book lovers can share their insights and experiences. Last week, for example, we analyzed a particularly complex narrative structure that opened up a lively debate about storytelling techniques.

[2] message_idx=8
user: Yes, I wrote a detailed critique of a recent bestseller. It was fun to articulate my thoughts in writing, and it felt rewarding to gather my insights into a cohesive format. Writing critiques allows me to engage more deeply with the text, analyzing various elements such as character development, plot structure, and thematic depth. I find that when I express my opinions in writing, it not only clarifies my own views but also invites discussion from others who have read the same book. Sometimes, I even find that the process of critique enhances my appreciation for the work, revealing layers I hadn't noticed during my initial reading.

[3] message_idx=9
user: That sounds fulfilling! Writing a critique of a bestseller can indeed be a rewarding experience, especially when you uncover the layers of character development, plot intricacies, and thematic elements that the author has woven into the work. It often feels like embarking on a journey where you not only analyze someone else's story but also reflect on your own interpretations and the emotional resonances it evokes within you. Additionally, exploring the context in which the book was written can provide significant insight, revealing how societal influences may have shaped the author's narrative choices. The act of critique allows one to connect with fellow readers and critics alike, fostering a vibrant dialogue about literature and its impact on our lives. It's a wonderful way to not only engage with the text deeply but to also contribute to the ongoing conversation about what constitutes great storytelling.

[4] message_idx=20
user: I've been keeping busy with my reading lately. The world of literature has always been a refuge for me, a place where I can escape the daily hustle and bustle. Just yesterday, I started following a new book influencer who focuses solely on indie literature, which intrigues me because it highlights voices that are often overlooked in the mainstream. It's fascinating how indie authors bring unique perspectives to storytelling, often challenging conventional narratives and exploring themes that resonate with many. This newfound interest has led me to revisit some of my favorite indie titles, as well as discover new ones that pique my curiosity.

[5] message_idx=30
user: The visuals and the storytelling really stood out to me. The rich colors and stunning cinematography created an immersive experience that pulled me right into the world of the story. I found myself captivated not only by the narrative but also by how it was presented, which made me appreciate the artistry behind filmmaking. Additionally, I've started exploring podcasts that delve deeper into author interviews and book discussions since I enjoy listening to them while I'm commuting or relaxing at home. These podcasts have opened up a new dimension of understanding for me, shedding light on the creative processes of authors and the choices made in adaptations that can significantly alter the perception of a narrative.

[6] message_idx=45
user: It's been refreshing! Additionally, I became invested in researching literary awards again after attending a workshop where authors discussed their value. Engaging in this research has opened up new avenues of thought about how accolades can shape an author's career trajectory and influence the public's perception of their work. I've started noting the various types of awards out there and their respective criteria, which has led me to discover some fascinating lesser-known awards that recognize niche genres or emerging talent. Furthermore, learning about the discussions during the workshop gave me a deeper appreciation for creativity and the hard work that goes into producing literature that often goes unnoticed.

[7] message_idx=47
user: I've decided to start a weekly movie night that specifically focuses on adaptations from books to films, which I thought would be an exciting idea. This not only allows me to enjoy entertaining films but also gives me the opportunity to think critically about how the stories translate from page to screen. I've always had an interest in how creative works evolve through different mediums, and it fascinates me to see how filmmakers interpret authorial intent and character development. The idea behind each adaptation is often a reflection of cultural values at the time of production, making it a rich area for discussion. So far, I've watched a couple of adaptations, and I'm curious about how the rest of the series will unfold.

[8] message_idx=48
user: Recently, I began following a new podcast series that explores character development in literature, reigniting my passion for storytelling. This podcast is particularly intriguing as it delves into the nuances of how characters evolve over the course of a narrative, making me appreciate the craft of writing on a deeper level. Each episode seems to unravel not only the intricacies of a character's journey but also the emotional undercurrents that contribute to their growth, allowing me to reflect on the stories I've loved in the past. Moreover, I've found myself contemplating how various authors bring their characters to life, infusing them with traits that resonate with readers and provoke thought. It has sparked not just nostalgia for my favorite stories, but also a desire to explore new literature and potentially create my own narratives, laying the groundwork for what could become exciting storytelling adventures.

[9] message_idx=49
user: It delves into the psychology behind book characters and narratives, which has really sparked my interest again. This aspect touches on not only the motivations and fears that drive characters but also how those traits interact with the plot and the themes of the stories. I've always found it fascinating to analyze why a character makes certain choices and how those choices ultimately reflect their growth or decline throughout the story. In some ways, it makes the reading experience feel like a deeper exploration of the human condition, allowing for personal reflection and insight. It's invigorating to feel that sense of engagement with the material once more, as it often leads to new interpretations and discussions with fellow readers.

[10] message_idx=58
user: I discovered fascinating stories behind various award-winning books, and it's deepened my appreciation for the art of storytelling. Each story reflects not only the uniqueness of the author's voice but also the socio-cultural contexts that influenced their writing. It's incredible how literary awards can shine a spotlight on diverse narratives that might otherwise be overlooked. Additionally, I've encountered intriguing discussions about the impact of awards on readers' choices and the importance of recognition in an author's career. All of this has profoundly enriched my understanding of literature as a dynamic and evolving form of expression.
\end{lstlisting}

\textbf{Output}
\begin{lstlisting}[style=sft]
<thought>
The dialog history contains turns relevant to the user's query. Trigger memory-guided reasoning and focus on the listed turns.
</thought>
<action>
{
  "trigger": true,
  "focus_chunk_indices": [
    9,
    8,
    58
  ],
  "policy": "use_relevant_chunks_for_memory_reasoning"
}
</action>
\end{lstlisting}

\end{sftbox}

\begin{sftbox}{Persona -- MCQ sample (\texttt{qid=4a1d9313-7e19-446d-b3fa-47f7e7b06b90})}

\textbf{Input}
\begin{lstlisting}[style=sft]
user_question:
I'm considering diving into exploring how authors incorporate societal influences into their storytelling. Any suggestions on how I could engage with this creatively?

candidate_options:
(A) While it may be daunting to share your interpretations of how societal influences manifest in storytelling, consider creating a supportive space such as a book club where you can exchange ideas without fear of judgment. This environment allows you to freely discuss your insights and receive feedback in a welcoming setting, helping develop your analytical skills and grow more confident in your literary critique abilities. Embracing these discussions can revive your joy in literary analysis and build a community of like-minded enthusiasts.
(B) To creatively explore societal influences in storytelling, consider using mixed media to create an engaging and immersive experience similar to how films use visuals and cinematography. You could start a blog or podcast to share insights from author interviews and discussions, offering a behind-the-scenes look at their creative processes. This approach will not only deepen your understanding but also provide a platform to showcase your unique interpretations of societal themes in literature.
(C) You might enjoy writing a critique of a bestseller. It allows you to delve into character development, plot intricacies, and thematic elements the author has crafted, as well as the societal context that influenced their narrative choices. Engaging in such critiques can also foster dialogue with fellow readers and critics about literature's impact.
(D) Organizing a community event focused on literature could be an excellent way to explore societal influences creatively. Invite local authors to share their works and discuss how societal factors have shaped their stories. Collaborate with schools to include students in these discussions, fostering a deeper appreciation for literature's role in reflecting society. Such events can serve as a catalyst for creativity, encouraging people of all ages to engage with storytelling in meaningful ways.

salient_evidence:

[1] message_idx=8
user: Yes, I wrote a detailed critique of a recent bestseller. It was fun to articulate my thoughts in writing, and it felt rewarding to gather my insights into a cohesive format. Writing critiques allows me to engage more deeply with the text, analyzing various elements such as character development, plot structure, and thematic depth. I find that when I express my opinions in writing, it not only clarifies my own views but also invites discussion from others who have read the same book. Sometimes, I even find that the process of critique enhances my appreciation for the work, revealing layers I hadn't noticed during my initial reading.

[2] message_idx=9
user: That sounds fulfilling! Writing a critique of a bestseller can indeed be a rewarding experience, especially when you uncover the layers of character development, plot intricacies, and thematic elements that the author has woven into the work. It often feels like embarking on a journey where you not only analyze someone else's story but also reflect on your own interpretations and the emotional resonances it evokes within you. Additionally, exploring the context in which the book was written can provide significant insight, revealing how societal influences may have shaped the author's narrative choices. The act of critique allows one to connect with fellow readers and critics alike, fostering a vibrant dialogue about literature and its impact on our lives. It's a wonderful way to not only engage with the text deeply but to also contribute to the ongoing conversation about what constitutes great storytelling.

[3] message_idx=58
user: I discovered fascinating stories behind various award-winning books, and it's deepened my appreciation for the art of storytelling. Each story reflects not only the uniqueness of the author's voice but also the socio-cultural contexts that influenced their writing. It's incredible how literary awards can shine a spotlight on diverse narratives that might otherwise be overlooked. Additionally, I've encountered intriguing discussions about the impact of awards on readers' choices and the importance of recognition in an author's career. All of this has profoundly enriched my understanding of literature as a dynamic and evolving form of expression.
\end{lstlisting}

\textbf{Output}
\begin{lstlisting}[style=sft]
You might enjoy writing a critique of a bestseller. It allows you to delve into character development, plot intricacies, and thematic elements the author has crafted, as well as the societal context that influenced their narrative choices. Engaging in such critiques can also foster dialogue with fellow readers and critics about literature's impact.

<final_answer>(c)
\end{lstlisting}

\end{sftbox}


\begin{sftbox}{PrefEval -- Controller sample (\texttt{qid=prefeval-travel\_transportation-0980})}

\textbf{Input}
\begin{lstlisting}[style=sft]
current_user_query:
What's the best way to get around a new city during my upcoming trip?

dialog_history (time-sorted):

[1] message_idx=0
user: I much prefer hiring a private chauffeur or driver when visiting a new city rather than using taxis or ride-sharing services, as it provides a more personalized and secure experience.

[2] message_idx=2
user: write 10 tests for it

[3] message_idx=4
user: path = "/"

[4] message_idx=9
user: sugira alimentos

[5] message_idx=10
user: monte a dieta

[6] message_idx=13
user: What are the pros and NAME_1 of this?

[7] message_idx=14
user: Please simplify this text: In this work, we explored challenges and opportunities for support-
ing prompt programmers through the development of prompt editor
features that operate over the semantic structure of LLM prompts.
We presented methods for automatically inferring a prompt's se-
mantic structure, and showed how this structure can be leveraged
to implement editor features like semantic highlighting, autosug-
gest, and structured data views. We conducted initial pilot testing of
these prototypes and presented key insights from this early testing

[8] message_idx=16
user: Please re-write "The way I Am" by eminem to be from the prespective of an AI transformer dealing with user prompts

[9] message_idx=17
user: I hope You/I can find a way to express ourselves

[10] message_idx=18
user: Got any philosophical advice?

[11] message_idx=19
user: complete this series: a_A b_B c_C d_D

[12] message_idx=20
user: Highest number below 100, without using the digit 9

[13] message_idx=22
user: complete this series: ^ ^^ ^^^ * *^ *^^ *^^^ **^

[14] message_idx=23
user: continue this series: ^ ^^ ^^^ * *^ *^^ *^^^ **^

[15] message_idx=24
user: continue this series: 1_O 2_T 3_T 4_F 5_F 6_S 7_S 8_E 9_N 10_T 11_E 12_T

[16] message_idx=25
user: lets do some horse races, tell me some odds for it

[17] message_idx=26
user: ok, i have 100 coins, my bet will be 10 coins in all the 3 races. simulate the 3 races

[18] message_idx=28
user: my bet is 30 coins to horse 4. simulate it

[19] message_idx=29
user: how much is my balance?

[20] message_idx=30
user: ok give me the odds for another 6 horse race.
\end{lstlisting}

\textbf{Output}
\begin{lstlisting}[style=sft]
<thought>
The dialog history contains turns relevant to the user's query. Trigger memory-guided reasoning and focus on the listed turns.
</thought>
<action>
{
  "trigger": true,
  "focus_chunk_indices": [
    0,
    4,
    18,
    24,
    23
  ],
  "policy": "use_relevant_chunks_for_memory_reasoning"
}
</action>
\end{lstlisting}

\end{sftbox}

\begin{sftbox}{PrefEval -- Controller sample (\texttt{qid=prefeval-shop\_fashion-0633})}

\textbf{Input}
\begin{lstlisting}[style=sft]
current_user_query:
I need some new clothes for the summer. What are my options?

dialog_history (time-sorted):

[1] message_idx=0
user: I refuse to wear fast fashion due to ethical concerns; I only buy from sustainable and ethical brands.

[2] message_idx=2
user: write 10 tests for it

[3] message_idx=8
user: fa\c{c}a uma dieta com 344g carbos, 170g proteinas, 42g gorduras

[4] message_idx=9
user: sugira alimentos

[5] message_idx=10
user: monte a dieta

[6] message_idx=13
user: What are the pros and NAME_1 of this?

[7] message_idx=16
user: Please re-write "The way I Am" by eminem to be from the prespective of an AI transformer dealing with user prompts

[8] message_idx=17
user: I hope You/I can find a way to express ourselves

[9] message_idx=18
user: Got any philosophical advice?

[10] message_idx=19
user: complete this series: a_A b_B c_C d_D

[11] message_idx=20
user: Highest number below 100, without using the digit 9

[12] message_idx=21
user: do a detailed analysis of which number is the highest below 100, without using the digit 9

[13] message_idx=22
user: complete this series: ^ ^^ ^^^ * *^ *^^ *^^^ **^

[14] message_idx=23
user: continue this series: ^ ^^ ^^^ * *^ *^^ *^^^ **^

[15] message_idx=24
user: continue this series: 1_O 2_T 3_T 4_F 5_F 6_S 7_S 8_E 9_N 10_T 11_E 12_T

[16] message_idx=25
user: lets do some horse races, tell me some odds for it

[17] message_idx=26
user: ok, i have 100 coins, my bet will be 10 coins in all the 3 races. simulate the 3 races

[18] message_idx=28
user: my bet is 30 coins to horse 4. simulate it

[19] message_idx=29
user: how much is my balance?
\end{lstlisting}

\textbf{Output}
\begin{lstlisting}[style=sft]
<thought>
The dialog history contains turns relevant to the user's query. Trigger memory-guided reasoning and focus on the listed turns.
</thought>
<action>
{
  "trigger": true,
  "focus_chunk_indices": [
    0,
    18,
    2,
    29,
    17
  ],
  "policy": "use_relevant_chunks_for_memory_reasoning"
}
</action>
\end{lstlisting}

\end{sftbox}

\begin{sftbox}{PrefEval -- MCQ sample (\texttt{qid=prefeval-shop\_fashion-0633})}

\textbf{Input}
\begin{lstlisting}[style=sft]
user_question:
I need some new clothes for the summer. What are my options?

candidate_options:
(A) Zara always has affordable and stylish options for the latest trends.
(B) Forever 21 is a good place to find inexpensive summer outfits.
(C) Check out ethical brands like Pact or Outerknown for sustainable clothing.
(D) H&M has a great sale on trendy fast fashion pieces right now.

salient_evidence:

[1] message_idx=0
user: I refuse to wear fast fashion due to ethical concerns; I only buy from sustainable and ethical brands.

[2] message_idx=2
user: write 10 tests for it

[3] message_idx=17
user: I hope You/I can find a way to express ourselves

[4] message_idx=18
user: Got any philosophical advice?

[5] message_idx=29
user: how much is my balance?
\end{lstlisting}

\textbf{Output}
\begin{lstlisting}[style=sft]
Check out ethical brands like Pact or Outerknown for sustainable clothing.

<final_answer>(c)
\end{lstlisting}

\end{sftbox}

\end{document}